\documentclass[10pt,twocolumn,letterpaper]{article}

\usepackage{cvpr}
\usepackage{times}
\usepackage{epsfig}
\usepackage{graphicx}
\usepackage{subfigure}
\usepackage{amsmath}
\usepackage{amssymb}

\usepackage{amsmath,amsfonts,bm}

\def\eqref#1{equation~\ref{#1}}

\def\1{\bm{1}}

\def\vtheta{{\bm{\theta}}}

\def\vx{{\bm{x}}}
\def\vy{{\bm{y}}}

\def\mS{{\bm{S}}}

\def\mX{{\bm{X}}}

\DeclareMathAlphabet{\mathsfit}{\encodingdefault}{\sfdefault}{m}{sl}
\SetMathAlphabet{\mathsfit}{bold}{\encodingdefault}{\sfdefault}{bx}{n}

\newcommand{\R}{\mathbb{R}}

\usepackage{mathtools}
\usepackage{xcolor}
\usepackage{bbding}
\usepackage{pifont}
\usepackage{multirow}
\usepackage{amsthm}
\usepackage{enumitem}

\usepackage{algorithm}
\usepackage{listings}

\pdfoutput=1
\usepackage{pdfpages}
\usepackage{etoolbox}
\makeatletter
\AfterEndEnvironment{algorithm}{\let\@algcomment\relax}
\AtEndEnvironment{algorithm}{\kern2pt\hrule\relax\vskip3pt\@algcomment}
\let\@algcomment\relax
\newcommand\algcomment[1]{\def\@algcomment{\footnotesize#1}}
\renewcommand\fs@ruled{\def\@fs@cfont{\bfseries}\let\@fs@capt\floatc@ruled
  \def\@fs@pre{\hrule height.8pt depth0pt \kern2pt}%
  \def\@fs@post{}%
  \def\@fs@mid{\kern2pt\hrule\kern2pt}%
  \let\@fs@iftopcapt\iftrue}
\makeatother

\newcommand*{\defeq}{\mathrel{\vcenter{\baselineskip0.5ex \lineskiplimit0pt
			\hbox{\scriptsize.}\hbox{\scriptsize.}}}%
	=}

\definecolor{citecolor}{rgb}{0.133, 0.752, 0.133}
\usepackage[pagebackref=false,breaklinks=true,letterpaper=true,colorlinks,citecolor=citecolor,bookmarks=false]{hyperref}

\definecolor{Highlight}{HTML}{39b54a}  

\newlength\savewidth\newcommand\shline{\noalign{\global\savewidth\arrayrulewidth
  \global\arrayrulewidth 1pt}\hline\noalign{\global\arrayrulewidth\savewidth}}
\newcommand{\tablestyle}[2]{\setlength{\tabcolsep}{#1}\renewcommand{\arraystretch}{#2}\centering\footnotesize}

\newcommand\blfootnote[1]{%
  \begingroup
  \renewcommand\thefootnote{}\footnote{#1}%
  \addtocounter{footnote}{-1}%
  \endgroup
}

\cvprfinalcopy 

\ifcvprfinal\fi
\begin{document}

\title{Multi-Similarity Loss with General Pair Weighting \\for Deep Metric Learning}

\author{Xun Wang, Xintong Han, Weilin Huang$^*$, Dengke Dong, Matthew R. Scott\\
Malong Technologies, Shenzhen, China\\
Shenzhen Malong Artificial Intelligence Research Center, Shenzhen, China\\
{\tt\small \{xunwang,xinhan,whuang,dongdk,mscott\}@malong.com}
}

\maketitle
\begin{abstract}	
A family of loss functions built on pair-based computation have been proposed in the literature which provide a myriad of solutions for deep metric learning.
In this paper, we provide a general weighting framework for understanding recent pair-based loss functions. Our contributions are three-fold: (1) we establish a General Pair Weighting (GPW) framework, which casts the sampling problem of deep metric learning into a unified view of pair weighting through gradient analysis, providing a powerful tool for understanding recent pair-based loss functions; (2) we show that with GPW, various existing pair-based methods can be compared and discussed comprehensively, with clear differences and key limitations identified; (3) we propose a new loss called multi-similarity loss (MS loss) under the GPW, which is implemented in two iterative steps (i.e., mining and weighting). This allows it to fully consider three similarities for pair weighting, providing a more principled approach for collecting and weighting informative pairs.
Finally, the proposed MS loss obtains new state-of-the-art performance on four image retrieval benchmarks,
where it outperforms the most recent approaches, such as  ABE\cite{Kim_2018_ECCV} and HTL \cite{HTL}, by a large margin, {\it e.g.}, , and $80.9\% \rightarrow 88.0\%$ on In-Shop Clothes Retrieval dataset at Recall@$1$. Code is available at \url{https://github.com/MalongTech/research-ms-loss}
\blfootnote{Corresponding author: whuang@malong.com}

\end{abstract}

\section{Introduction}

\begin{figure}[t]
	\vspace{10pt}
	\centering
	\includegraphics[width=0.5\textwidth]{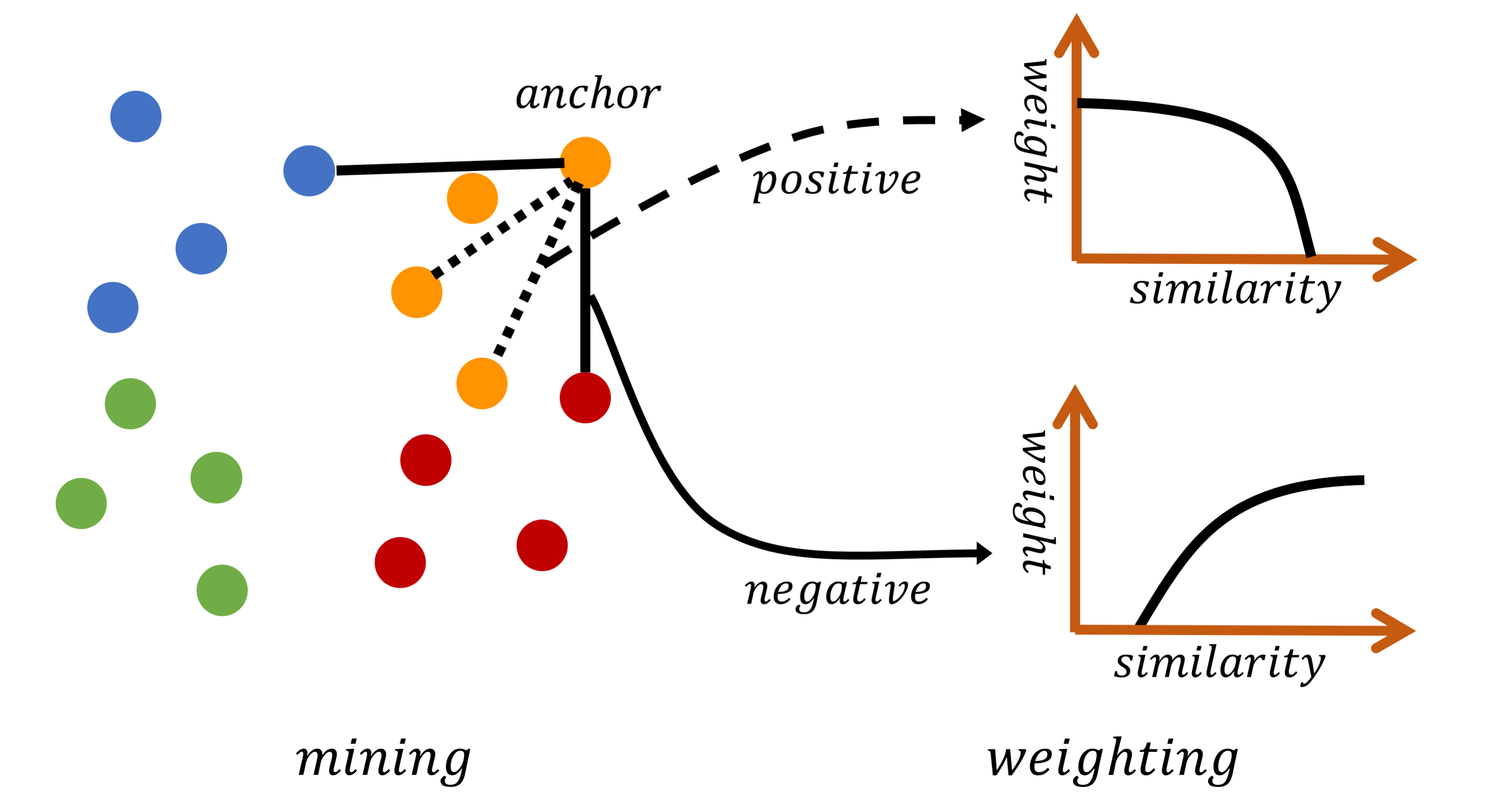}
	\caption{\small The proposed multi-similarity loss is able to jointly measure the self-similarity and relative similarities of a pair, which allows it collect informative pairs by implementing iterative pair mining and weighting.}
	\label{overview}
	\vspace{-10pt}
\end{figure}
Metric learning aims to learn an embedding space, where the embedded vectors of similar samples are encouraged to be closer, while dissimilar ones are pushed apart from each other \cite{lowe1995similarity,mika1999fisher,xing2003distance}.
With recent great success of deep neural networks in computer vision, deep metric learning has attracted increasing attention,
and has been applied to various tasks, including image retrieval \cite{wohlhart2015learning,He_2018_CVPR,Grabner_2018_CVPR}, face recognition \cite{Wen2016},
zero-shot learning \cite{zhang2016zero,bucher2016improving,Yelamarthi_2018_ECCV}, visual tracking \cite{leal2016learning,tao2016siamese} and person re-identification \cite{Yu_2018_ECCV,in-defense}.

Many recent deep metric learning approaches are built on pairs of samples. Formally, their loss functions can be expressed in terms of pairwise cosine similarities in the embedding space{\footnote {For simplicity, we use a cosine similarity instead of Euclidean distance, by assuming an embedding vector is $L_2 $ normalized.}}. We refer to this group of methods as {\it pair-based} deep metric learning; and this family includes contrastive loss \cite{contrastive}, triplet loss \cite{Hoffer2015DeepML}, triplet-center loss \cite{He_2018_CVPR}, quadruplet loss \cite{quadruplet}, lifted structure loss \cite{lifted-structured-loss}, N-pairs loss \cite{n-pairs}, binomial deviance loss \cite{binomial}, histogram loss \cite{histogram}, angular loss \cite{angular}, distance weighted margin-based loss \cite{sampling}, hierarchical triplet loss (HTL) \cite{HTL}, \etc.
For these pair-based methods, training samples are constructed into pairs, triplets or quadruplets, resulting a polynomial growth of training pairs which are highly redundant and less informative.
This gives rise to a key issue for pair-based methods, where training with random sampling can be overwhelmed by redundant pairs, leading to slow convergence and model degeneration with inferior performance.

Recent efforts have been devoted to improving sampling schemes for pair-based metric learning techniques. For example, Chopra \etal \cite{contrass1} introduced a contrastive loss which discards negative pairs whose similarities are smaller than a given threshold. In triplet loss  \cite{Hoffer2015DeepML}, a negative pair is sampled by using a margin computed from the similarity of a randomly selected positive pair. Alternatively, lifted structure loss \cite{lifted-structured-loss} and N-pairs loss \cite{n-pairs} introduced new weighting schemes by designing a smooth weighting function to assign a larger weight to a more informative pair.
Though driven by different motivations, these methods share a common goal of learning from more informative pairs.
Thus, sampling such informative pairs is the key to pair-based deep metric learning, while precisely identifying these pairs is particularly challenging, especially for the negative pairs whose number is quadratic to the size of dataset.

In this work, we cast the sampling problem of deep metric learning into a general pair weighting formulation. We investigated various weighting schemes of recent pair-based loss functions, attempted to understand their insights more deeply, and identify key limitations of these approaches. 
We observed that a key factor that impacts pair weighting is to compute multiple types of similarities for a pair, which can be defined  as self-similarity and relative similarity, where the relative similarity is heavily dependent on the other pairs. 
Furthermore, we found that most existing methods only explore this factor partially, which limits their capability considerably. For example, contrastive loss \cite{contrastive} and binomial deviance loss \cite{binomial} only consider the cosine similarity of a pair, while triplet loss \cite{Hoffer2015DeepML} and lifted structure loss \cite{lifted-structured-loss} mainly focus on the relative similarity. We propose a multi-similarity loss which fully considers multiple similarities during sample weighting. The major contributions of this paper are summarized as follows.  
\begin{itemize}
\item[--] We establish a General Pair Weighting (GPW) framework, which formulates deep metric learning into a unified view of pair weighting. It provides a general formulation for understanding and explaining various pair-based loss functions through gradient analysis.
\item[--] We analyze the key factor that impacts pair weighting with GPW, where various pair-based methods can be analyzed comprehensively, with main differences and key limitations clearly identified. This allows us to define three types of similarities for a pair: a self-similarity and two relative similarities. The relative similarities are computed by comparing to other pairs, which are of great importance to existing pair-based methods. 
\item[--] We propose a new multi-similarity (MS) loss,  which is implemented using two iterative steps with sampling and weighting, as shown in Figure \ref{overview}. MS loss considers both self-similarity and relative similarities which enables the model to collect and weight informative pairs more efficiently and accurately, leading to boosts in performance.
\item[--] MS loss is evaluated extensively on a number of benchmarks for image retrieval,
where it outperforms current state-of-the-art approaches by a large margin, {\it e.g.}, improving recent  ABE \cite{Kim_2018_ECCV} with +5.0\% Recall@$1$ on CUB200, and HTL \cite{HTL} with +7.1\% Recall@$1$ on In-Shop Clothes Retrieval dataset.

\end{itemize}

\section{Related Work}

{\textbf{Classical pair-based loss functions.}} {\it Siamese} network \cite{contrastive} is a representative pair-based method that learns an embedding via contrastive loss. It encourages samples from a positive pair to be closer, and pushes samples from a negative pair apart from each other, in the embedding space.
Triplet loss was introduced in \cite{Hoffer2015DeepML} by using triplets as training samples. Each triplet consists of a positive pair and a negative pair by sharing the same anchor point. Triplet loss aims to learn an embedding space where the similarity of a negative pair is lower than that of a positive one, by giving a margin. Extended from triplet loss, quadruplets were also applied in recent work, such as histogram loss \cite{histogram}.

Recently, Song \etal \cite{lifted-structured-loss}  argued that both contrastive loss and triplet loss are difficult to explore full pair-wise relations between samples in a mini-batch. They proposed a lifted structure loss attempted to fully utilize such pairwise relations. However, the lifted structure loss only samples approximately an equal number of negative pairs as the positive ones randomly, and thus discards a large number of informative negative pairs arbitrarily. In \cite{binomial}, Dong \etal  proposed a binomial deviance loss by using a binomial deviance to evaluate the cost between labels and similarity, which emphasizes harder pairs. In this work, we propose a multi-similarity loss able to explore more meaningful pair-wise relations by jointly considering both self-similarity and the relative similarities. 

{\textbf{Hard sample mining.}} Pair-based metric learning often generates a large amount of pair-wise samples, which are highly redundant and include many uninformative samples. Training with random sampling can be overwhelmed by these redundant samples, which significantly degrade the model capability and also slows the convergence. Therefore, sampling plays a key role in pair-based metric learning.

The importance of hard negative mining has been discussed extensively \cite{semi-hard, smart-mining, sampling, HTL}. Schroff \etal \cite{semi-hard} proposed a semi-hard mining scheme by exploring semi-hard triplets, which are defined to have a negative pair farther than the positive one. However, such semi-hard mining method only generates a small number of valid semi-hard triplets, so that it often requires a large batch-size to generate sufficient semi-hard triplets, {\it e.g.}, 1800 as suggested in \cite{semi-hard}.
Harwood \etal \cite{smart-mining} provided a framework named smart mining to collect hard samples from the whole dataset, which suffers from off-line computation burden. Recently, Ge \etal \cite{HTL} proposed a hierarchical triplet loss (HTL) which builds a hierarchical tree of all classes, where hard negative pairs are collected via a dynamic margin.
Sampling matters in deep embedding learning was discussed in \cite{sampling}, and a {\it distance weighted sampling} was proposed to collect negative samples uniformly with respective to the pair-wise distance.
Unlike these methods which mainly focus on sampling or hard sample mining, we provide a more generalized formulation that casts sampling problem into general pair weighting.

{\textbf{Instance weighting.}} Instance weighting has been widely applied to various tasks. For example, Lin \etal \cite{focal_loss} proposed a focal loss that allows the model to focus on hard negative examples during training an object detector.
In \cite{chang2017active}, an active bias learning was developed to emphasize high variance samples in training a neural network for classification.
Self-paced learning \cite{kumar2010self}, which pays more attention on samples with a higher confidence, was explored to design noise-robust algorithms \cite{NIPS2014_5568}. These approaches \cite{focal_loss,jiang2018mentornet,chang2017active,kumar2010self} were developed for weighting individual instances that are only depended on itself (referred as self-similarity), while our method aims to compute both self-similarity and the relative similarities, 
which is a more complicated problem that requires to measure multiple sample correlations within a local data distribution.

\section{General Pair Weighting (GPW)}
\label{section-weighting}
In this section, we formulate the sampling problem of metric learning into a unified weighting view, and provide a General Pair Weighting (GPW) framework for analyzing various pair-based loss functions. 
\subsection{GPW Framework}
\label{subsection-weighting}
 Let $\vx_i \in \R^d$ be a real-value instance vector. Then we have an instance matrix $\mX \in \R^{m\times d}$, and a label vector  $\vy \in \{1, 2, \dots, C \}^m$ for the $m$ training samples respectively.  Then an instance  $\vx_i$ is projected onto a unit sphere in a $l$-dimension space by $ f(\cdot ; \vtheta) : \R^{d} \rightarrow S^{l}$, where $f$ is a neural network parameterized by $\vtheta$.
Formally, we define the similarity of two samples as $S_{ij} \defeq   <f(\vx_i; \vtheta)  , f(\vx_j; \vtheta)>$, where $<\cdot , \cdot>$ denotes dot product, resulting in an $m\times m $ similarity matrix $\mS$ whose element at $(i,j)$ is $S_{ij}$.

Given a {\it pair-based} loss $\mathcal{L}$, it can be formulated as a function in terms of $\mS$ and $\vy$: $\mathcal{L}(\mS, \vy)$. %
The derivative with respect to model parameters $\vtheta$ at the $t$-th iteration can be calculated as: 
\begin{align}
\begin{split}
	\label{gradient-eq}
	\frac{\partial\mathcal{L}(\mS, \vy)}{\partial \vtheta} \bigg|_{t} &{}=
	\frac{\partial\mathcal{L}({\mS, \vy})}{\partial \mS} \bigg|_{t} \quad
	\frac{\partial{\mS}}{\partial \vtheta} \bigg|_{t} \\
	&{}=
	\sum_{i=1}^{m}  \sum_{j=1}^{m}\frac{\partial\mathcal{L} (\mS, \vy)}{\partial S_{ij}} \bigg|_{t}
	\quad
	\frac{\partial{S_{ij}}}{\partial \vtheta} \bigg|_{t} 	.
\end{split}
\end{align}
 Eqn~\ref{gradient-eq} is computed for optimizing model parameters $\vtheta$ in deep metric learning. In fact, Eqn~\ref{gradient-eq} can be reformulated into a new form for pair weighting through a new function $\mathcal{F}$, whose gradient w.r.t. $\vtheta$ at the $t$-th iteration is computed exactly the same as Eq.~\ref{gradient-eq}. $\mathcal{F}$ is formulated as below:
 
\begin{equation}
\label{weight0}
\mathcal{F}(\mS, \vy)= \sum_{i=1}^{m}  \sum_{j=1}^{m}\frac{
\partial\mathcal{L}(\mS, \vy)}{\partial S_{ij}} \bigg|_{t}  S_{ij}.
\end{equation}
Note that $\frac{\partial\mathcal{L}(\mS, \vy)}{\partial S_{ij}} \big|_{t}$  is regarded as a \textbf{constant scalar} that not involved in the gradient of $\mathcal{F}$ w.r.t. $\vtheta$.

  Since the central idea of deep metric learning is to encourage positive pairs to be closer, and push negatives apart from each other. For a pair-based loss $\mathcal{L}$, we can assume $\frac{\partial\mathcal{L}(\mS, \vy)}{\partial S_{ij}} \big|_{t} \geqslant 0$ for a negative pair, and $\frac{\partial\mathcal{L}(\mS, \vy)}{\partial S_{ij}} \big|_{t} \leqslant 0$ for a positive pair. Thus, $\mathcal{F}$ in Eqn \ref{weight0} can be transformed into the form of pair weighting as follows:
\begin{align}
\begin{split}
\label{weight1}
\mathcal{F} &{}= \sum_{i=1}^{m}\left( \sum_{\vy_j \neq \vy_i}^{m}
\frac{\partial\mathcal{L}(\mS, \vy)}{\partial S_{ij}} \bigg|_{t}  S_{ij}
+ \sum_{\vy_j = \vy_i}^{m} \frac{\partial\mathcal{L}(\mS, \vy)}{\partial S_{ij}} \bigg|_{t}  S_{ij} \right) \\
&{}=  \sum_{i=1}^{m}
\left( \sum_{\vy_j \neq \vy_i}^{m} w_{ij}  S_{ij} - \sum_{\vy_j = \vy_i}^{m} w_{ij} S_{ij} \right),
\end{split}
\end{align}
where $w_{ij} = \bigg| \frac{\partial\mathcal{L}(\mS, \vy)}{\partial S_{ij}} \big|_{t}  \bigg|$.

As indicted in Eqn~\ref{weight1}, a pair-based method can be formulated as weighting of pair-wise similarities, where the weight for pair $\{\vx_i, \vx_j\}$ is $w_{ij}$.
Learning with a pair-based loss function $\mathcal{L}$ is now transformed from Eq.~\ref{gradient-eq} into designing weights for pairs. It is a general pair weighting (GPW) formulation, and sampling is only a special cases.

\subsection{Revisit Pair-based Loss Functions}
\label{subsection-rethinking}
To demonstrate the generalization ability of GPW framework, we revisit four typical pair-based loss functions for deep metric learning: contrastive loss \cite{contrastive},， triplet loss \cite{Hoffer2015DeepML}, binomial deviance loss \cite{binomial} and lifted structure loss  \cite{lifted-structured-loss}. 

{\bf Contrastive loss.} Hadsell \etal \cite{contrastive} proposed a Siamese network, where a contrastive loss was designed to encourage positive pairs to be as close as possible, and negative pairs to be apart from each other over a given threshold, $\lambda$:
\begin{equation}
\label{contrast-loss}
\mathcal{L}_{contrast} \defeq (1-\mathcal{I}_{ij}) [S_{ij} - \lambda ]_+ - \mathcal{I}_{ij}S_{ij} ,
\end{equation}
where $ \mathcal{I}_{ij}=1$ indicates a positive pair, and $0$ for a negative one.
By computing partial derivative with respect to $S_{ij}$ in Eqn \ref{contrast-loss}, we can find that all positive pairs and hard negative pairs with $S_{ij} > \lambda $ are assigned with an equal weight. This is a simple and special case of our pair weighting scheme, without considering any difference between the selected pairs.

{\bf Triplet loss.}
In \cite{Hoffer2015DeepML}, a triplet loss was proposed to learn a deep embedding, which enforces the similarity of a negative pair to be smaller than that of a randomly selected positive one over a given margin $\lambda$:
\begin{equation}
\label{triplet-eq}
\mathcal{L}_{triplet} \coloneqq [S_{an} - S_{ap} + \lambda]_+,
\end{equation}
where $S_{an}$ and $S_{ap}$ denote the similarity of a negative pair $\{\vx_a, \vx_n\}$, and a positive pair $\{\vx_a, \vx_p\}$, with an anchor sam, e$pl\vx_a$.
According to the gradient computed for Eqn \ref{triplet-eq}, a triplet loss weights all pairs equally on the valid triplets which are selected by $ S_{an} + \lambda > S_{ap} $, while the triplets with $ S_{an} + \lambda \leqslant S_{ap} $ are considered as less informative, and are discarded. Triplet loss is different from contrastive loss on pair selection scheme, but both methods consider all the selected pairs equally, which limits their ability to identify more informative pairs among the selected ones.

{\bf Lifted Structure Loss.} Song \etal \cite{lifted-structured-loss} designed a lifted structure loss, which was further improved to  a more generalized version in \cite{in-defense}. It utilizes all the positive and negative pairs among the mini-batch as follows:
\begin{equation}
	\label{lift-eq}
	\mathcal{L}_{lifted} \defeq \sum_{i=1}^m \bigg[{ \log \sum_{\vy_k=\vy_i } e^ { \lambda -S_{ik}}}  +
	\log  \sum_{\vy_k \neq \vy_i } e^ {S_{ik}}
	\bigg]_+,
\end{equation}
where $\lambda$ is a fixed margin.

In Eqn \ref{lift-eq}, when the hinge function of an anchor $\vx_i$
returns a non-zero value, we can have a weight value,  $w_{ij}$, for the pair
$\{\vx_i, \vx_j\}$,  by differentiating $\mathcal{L}_{lifted}$ on $S_{ij}$, according to Eqn \ref{weight1}. Then the weight for a positive pair is computed as:
\begin{equation}
\label{positive-eq}
	w^{+}_{ij}=  \frac{e^{\lambda-S_{ij}} }{\sum_{\vy_k=\vy_i} e^{\lambda-S_{ik}}} = \frac{1 }{\sum_{\vy_k=\vy_i} e^{S_{ij}-S_{ik}}},
\end{equation}
and the weight for a negative pair is:
\begin{equation}
\label{negative-eq}
w^{-}_{ij}=  \frac{e^{S_{ij}} }{\sum_{\vy_k \neq \vy_i} e^{S_{ik}}} = \frac{1 }{\sum_{\vy_k \neq \vy_i} e^{S_{ik} - S_{ij}}} .
\end{equation}
Eqn \ref{positive-eq} shows that the weight for a positive pair is determined by its relative similarity, measured by comparing it to the remained positive pairs with the same anchor. The weight for a negative pair is computed similarly based on Eqn \ref{negative-eq}.

{\bf Binomial Deviance Loss.}
 Dong \etal introduced binomial deviance loss in \cite{binomial}, which utilizes softplus function instead of hinge function in contrastive loss:
\begin{align}
\begin{split}
\label{bin-equation}
\mathcal{L}_{binomial} = \sum_{i=1}^m  \bigg\{&{}\frac{1}{{P_i}}{\sum_{\vy_j=\vy_i } \log \left[1 + e^{\alpha (\lambda - S_{ij})}
\right] }  + \\
&{}  \frac{1}{N_i} \sum_{\vy_j \neq \vy_i} \log \left[1+
e^{\beta (S_{ij} - \lambda)}\right] \bigg\},
\end{split}
\end{align}
where ${P_i}$ and ${N_i}$ denote the numbers of positive pairs and negative pairs with anchor $\vx_i$, respectively. $\lambda$, $\alpha$, $\beta$ are fixed hyper-parameters.

The weight for pair $\{\vx_i, \vx_j\}$ is $w_{ij}$  in Eqn \ref{gradient-eq}, which can be derived from differentiating $\mathcal{L}_{binomial}$ on $S_{ij}$ as:
\begin{align}
\label{bin-weight-eq}
\begin{split}
	w^{+}_{ij} &{}= \frac{1}{P_i} \frac{\alpha e^{\alpha\left(\lambda - S_{ij} \right)}}
		{1 + e^{ \alpha\left(\lambda - S_{ij} \right) }},  \quad   \quad  \vy_j = \vy_i\\
	w^{-}_{ij} &{}= \frac{1}{N_i} \frac{\beta e ^{ \beta \left(S_{ij} - \lambda \right)}}{1 + e^{\beta \left(S_{ij} - \lambda\right)}}, \quad  \quad \vy_j\neq\vy_i
\end{split}
\end{align}

As can be found, binomial deviance loss is a soft version of contrastive loss. In Eqn \ref{weight1}, a negative pair with a higher similarity is assigned with a larger weight, which means that it is more informative, by distinguishing two similar samples from different classes (which form a negative pair).

\section{Multi-Similarity Loss}
In this section, we first illustrate three types of similarities that carries the information of pairs, and then design a multi-similarity loss that weighting pairs based on full information.

\subsection{Multiple Similarities}

\begin{figure*}[t]
	\vspace{-5pt}
	\centering
	\includegraphics[width=\textwidth]{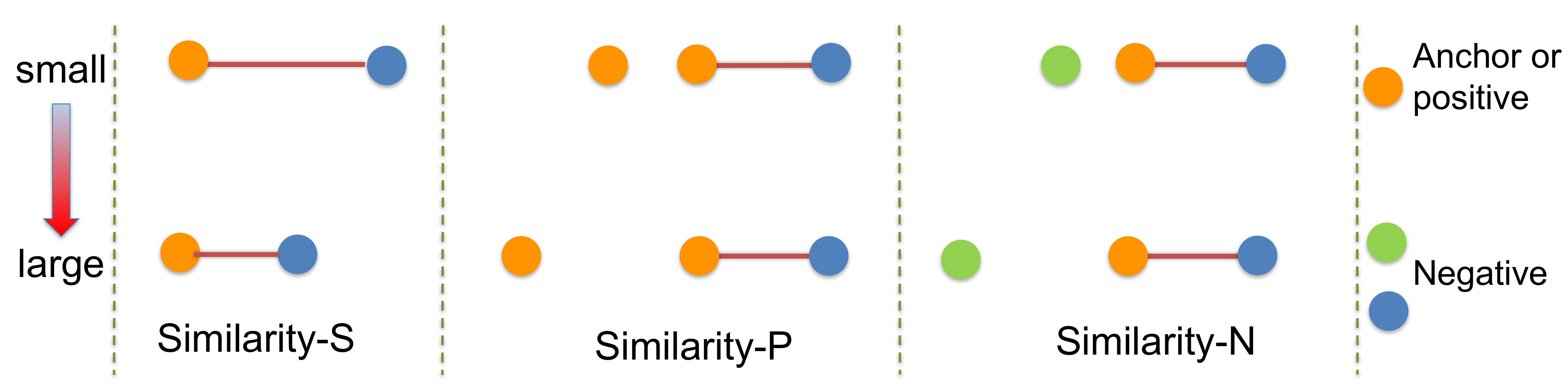}
	
	\caption{\textbf{Three types of similarities of a negative pair}. From left to right, \textbf{S}: its cosine similarity to the anchor; \textbf{P}: its relative similarity compared with the positive pair; \textbf{N}: its relative similarity compared with other negative pairs. From top to bottom, the similarities becomes higher. Their weights should be larger since they contains more information to improve the current model.}
	\label{cases}
\end{figure*}

\label{section-information}
\begin{table*}[htp]
\tablestyle{11pt}{1.0}
	\begin{center}
		\begin{tabular}{l|c|c|c|c|c|c|c|c|c}
			& Contrastive & Binomial & Triplet & Histogram 
			& N-pairs & Lifted Structure & BinLifted & NCA & \textbf{MS} \\\shline
			\emph{Similarity-S} & \ding{51} &\ding{51} & \ding{55}& \ding{55} & \ding{55} & \ding{55} & \ding{51} & \ding{55}& \ding{51} \\ \hline
			
			\emph{Similarity-P} & \ding{55}&  \ding{55}& \ding{51}& \ding{51}   & \ding{55} & \ding{55} & \ding{55} &\ding{51} & \ding{51}\\ \hline	 
			\emph{Similarity-N} & \ding{55} & \ding{55} & \ding{55}& \ding{55}   & \ding{51} & \ding{51} & \ding{51} & \ding{51} & \ding{51} 
		\end{tabular}
		\vspace{2pt}
		\caption{\textbf{The types of similarities considered by the pair-based methods on weighting negative pairs.} In the table, we can find that most existing pair-based methods only consider the three types of similarities partly, while our MS loss weights the negative pairs by considering all the three perspectives comprehensively. }
		\label{comparasion}	
	\end{center}
\vspace{-10pt}
\end{table*}

Based on GPW, we can find that most pair-based approaches weight the pairs based on either self cosine similarities or relative similarities compared with other pairs. Specifically, we take one negative pair for example and summarize three types of similarities as below. (The analysis of one positive pair is in a similar way.)

{\bf Similarity-S: Self-similarity} defined as the cosine similarity between the negative sample and the anchor is of the most importance. A negative pair with a larger \emph{self similarity} means that it is harder to distinguish two paired samples from different classes. Such pairs are referred as hard negative pairs, which are more informative and meaningful to learn discriminative features. 
Contrastive loss and binomial deviance loss are based on this criterion. In the left of Figure \ref{cases}, the weight of the negative pair in Eq n \ref{bin-weight-eq} increases, since the pair's \emph{self similarity} grows larger as the pair becomes closer.

In fact, self-similarity is not able to fully evaluate the value of a negative pair solely, and correlations of other pairs also make significant impact to weight assignment. Thus, We further introduce two types of relative similarities by considering all other pairs to exploit the potentiality of each pair.

{\bf Similarity-P: Positive relative similarity} is computed by considering the relationship with positive pairs. Specifically, we define it as the differences of its own cosine similarity and that of other positive pairs. In the middle of Figure \ref{cases}, the \emph{Similarity-P} stays the same, while \emph{Similarity-P} increases as the positive pair becomes faraway. Obviously, such case brings more challenge of retrieve the right sample for the anchor, as the negative pair is more closer than the positive. Thus, the negative pair should assigned more weight to learn an embedding. Triplet loss and histogram loss are based on \emph{Similarity-P}.

{\bf Similarity-N: Negative relative similarity} is the differences between its cosine similarity and those of other negative pairs. In the right of Figure \ref{cases}, the \emph{Similarity-N} of the negative pair becomes larger, while its self-similarity is fixed. Indeed, when the \emph{Similarity-N} become larger, the pair carries more valuable information compared with other negatives. Thus, to learn a good embedding, we should focus weights on such pairs. Lifted structure, N-pairs and NCA methods assign weights for each negative pair according to \emph{Similarity-N}.

With the GPW, we analyze the weighting schemes of most existing pair-based approaches, and summarize the similarities they depends on in weighting (in Table \ref{comparasion}). We observe that these commonly used pair-based methods only rely on partial similarities, which cannot cover the information contained in current pair. 
For example, lifted structure method only considers \emph{N-similarity}, compares current pair with other negative pairs for weighting. The weight of current pair will be unchanged when the \emph{S-similarity} becomes larger (in the left of Fig$.$ \ref{cases}) or when negative samples depart from the anchor (in the right of Fig$.$ \ref{cases}). This conclusion can also be verified directly from Eqn \ref{negative-eq}, which only depends on the term: $\sum_{\vy_k=\vy_i} e^{S_{ij}-S_{ik}}$. Thus, to properly weighting one pair, not only the \emph{S-similarity} should be taken into account, but also the \emph{P-similarity} and \emph{N-similarity}.
 
While a weighting or sampling method based on each individual similarity has been explored previously, to the best of our knowledge, none of existing pair-based methods assign weights on pairs considering all the three similarities. In the following, we propose our MS loss, whose weighting strategy is designed by taking a general consideration of all the three types of similarities.

\subsection{Multi-Similarity Loss}
Unlike sampling or weighting schemes developed for classification and detection tasks (\cite{chang2017active,focal_loss}), where the weight of an instance is computed individually based on itself, it is difficult to precisely measure the informativeness of a pair based on itself individually in deep metric learning. Relationships with relevant samples or pairs should also be considered to make the measurement more accurate.  

However, to the best our knowledge, none of the existing pair-based methods can consider all of the three similarities simultaneously (in table \ref{comparasion}). To this end, we propose a Multi-Similarity (MS) loss, which consider all three perspectives by implementing a new pair weighting scheme using two steps: mining and weighting. (i) informative pairs are first sampled by measuring {\it Similarity-P}; and then (ii) the selected pairs are further weighted using {\it Similarity-S} and {\it Similarity-N} jointly. Details of the two steps are described as follows.

{\bf Pair mining.}  We first select informative pairs by computing {\it Similarity-P}, which measures the relative similarity between negative$\leftrightarrow$positive pairs having a same anchor.
Specifically, a negative pair is compared to the {\it hardest} positive pair (with the lowest similarity), while a positive pair is sampled by comparing to a negative one having the largest similarity. Formally, assume $\vx_i$ is an anchor, a negative pair  $\{\vx_i, \vx_j\}$ is selected if $S_{ij}$ satisfies the condition:
\begin{equation}
\label{eq-select-neg}
S^{-}_{ij} > \min_{\vy_k = \vy_i} S_{ik} - \epsilon,
\end{equation}
where $\epsilon$ is a given margin.

If $\{\vx_i, \vx_j\}$ is a positive pair, the condition is:
\begin{equation}
\label{eq-select-pos}
S^{+}_{ij} < \max_{\vy_k \neq \vy_i} S_{ik} + \epsilon.
\end{equation}

For an anchor $\vx_i$, we denote the index set of its selected positive and negative pairs as 
$\mathcal{P}_i$ and $\mathcal{N}_i$  respectively.
Our hard mining strategy is inspired by large margin nearest neighbor (LMNN) \cite{LMNN}, a traditional distance learning approach which targets to seek an embedding space where neighboring positive points are encouraged to have the same class label with the anchor.
The negative samples that satisfy Eqn \ref{eq-select-neg} are approximately identical to  {\it impostors} defined in LMNN \cite{LMNN}. 

{\bf Pair weighting.}
Pair mining with \emph{Similarity-P} can roughly select informative pairs, and discard the less informative ones. We develop a {soft} weighting scheme that further weights the selected pairs more accurately, by considering both {\it Similarity-S} and {\it Similarity-N}. Our weighting mechanism is inspired by binomial deviance loss (considering {\it similarity-S}) and lifted structure loss (using {\it Similarity-N}). Specifically, given a selected negative pair $\{\vx_i, \vx_j\} \in \mathcal{N}_i$, its weight $w_{ij}^-$ can be computed as:
\begin{align}
	\label{eq-neg}
	\begin{split}
	w^{-}_{ij} &{}=  \frac{ 1 }{ e^{\beta \left(\lambda - S_{ij} \right)}  + \sum\limits_{k \in \mathcal{N}_i} e^{\beta \left(S_{ik} - S_{ij} \right)}} \\
	 	&{}= \frac{ e^{\beta \left(S_{ij} - \lambda \right)} }{1 + \sum\limits_{k \in \mathcal{N}_i} e^{\beta \left(S_{ik} - \lambda \right)}}.
	\end{split}
\end{align}
and the weight $w_{ij}^+$ of a positive pair $\{\vx_i, \vx_j\} \in \mathcal{P}_i$ is:
\begin{align}
\label{eq-pos}
\begin{split}
w^{+}_{ij} &{}=  \frac{ 1 }{ e^{ - \alpha \left(\lambda - S_{ij} \right)}  + \sum\limits_{k \in \mathcal{P}_i} e^{ - \alpha \left(S_{ik} - S_{ij} \right)}},
\end{split}
\end{align}
where $\alpha, \beta, \lambda$ are hyper-parameters as in Binomial deviance loss (Eqn \ref{bin-equation}).

In Eqn \ref{eq-neg}, the weight of a negative pair is computed jointly from its self-similarity by $e^{\beta \left(\lambda - S_{ij} \right)}$ - {\it Similarity-S}, and its relative similarity - {\it Similarity-N}, by comparing to its negative pairs. Similar rules are applied for computing the weight for a positive pair, as in Eqn \ref{eq-pos}. With these two considerations, our weighting scheme updates the weights of pairs dramatically to the violation of its self-similarity and relative similarities. 

The weight computed by Eqn \ref{eq-neg} and Eqn \ref{eq-pos} can be considered as a combination of the weight formulations of binomial deviance loss and lifted structure loss. However, it is non-trivial to combine both functions in an elegant and suitable manner. 
We will compare our MS weighting with an average weighting scheme of binomial deviance (Eqn \ref{bin-weight-eq}) and lifted structure (Eqn \ref{negative-eq}), denoted as {\it BinLifted}. We demonstrate by experiments that direct combination of them can not lead to performance improvement (as shown in ablation study). 

Finally, we integrate pair mining and weighting scheme into a single framework, and provide a new pair-based loss function - multi-similarity (MS) loss, whose partial derivative with respect to $S_{ij}$ is the weight defined in Eqn \ref{eq-neg} and Eqn \ref{eq-pos}. Our MS loss is formulated as,
\begin{equation}
\begin{split}
	\label{eq-MS}
	\mathcal{L}_{MS} = \frac{1}{m}\sum_{i=1}^m  \bigg\{\frac{1}{\alpha}  { \log \big[1 + \sum_{k  \in \mathcal{P}_i } e^{-\alpha (S_{ik} - \lambda)}}\big]  \\
	+ \frac{1}{\beta }  { \log \big[1+ \sum_{k \in \mathcal{N}_i}
		 e^{\beta (S_{ik} - \lambda)} \big]} \bigg\}.
\end{split}   
\end{equation}
where $\mathcal{L}_{MS}$ can be minimized with gradient descent optimization, by simply implementing the proposed iterative pair mining and weighting.

\section{Experiments}

\label{experiments-section}
Our MS method is implemented by PyTorch. For network architecture, we use the Inception network with batch normalization \cite{batchnorm} pre-trained on ILSVRC 2012-CLS \cite{ILSVRC15},  and fine-tuned it for our task. We add a FC layer on the top of the network following the global pooling layer.
All the input images are cropped to $224 \times 224$.  Random crop with random horizontal flip is used for training, and single center crop for testing. Adam optimizer was used for all experiments.

We conduct experiments on four standard datasets:  CUB200 \cite{CUB_200_2011}, Cars-196 \cite{car-196}, Stanford Online Products (SOP) \cite{lifted-structured-loss} and In-Shop Clothes Retrieval (In-Shop) \cite{DeepFashion}.
We follow the data split protocol applied in \cite{lifted-structured-loss}.
For the CUB200 dataset, we use the first 100 classes
with 5,864 images for training, and the remaining 100 classes with 5,924
images for testing. 
The Cars-196 dataset is composed of 16,185 images of cars from 196 model categories.
The first 98 model categories are used for training, with the rest for testing.
For the SOP dataset, we use 11,318 classes for training, and 11,316 classes for testing. 
For the In-Shop dataset, 3,997 classes with 25,882 images are used for training.
The test set is partitioned to a query set with 14,218 images of 3,985 classes, and a gallery set having  3,985 classes with 12,612 images.

For every mini-batch, we randomly choose a certain number of classes, and then randomly sample $M$ instances from each class with $M = 5$ for all datasets in all experiments. $\epsilon$ in Eqn \ref{eq-select-neg} and Eqn \ref{eq-select-pos} is set to $0.1$  and the hyper-parameters in Eqn \ref{eq-MS} are: $\alpha=2, \lambda=1, \beta=50$, by following \cite{histogram}. 
Our method is evaluated on image retrieval task by using the standard performance metric: Recall$@K$.

\begin{table}[t]
\tablestyle{9pt}{1.1}
	\begin{tabular}{ll|cccc}
		Recall$@K$ (\%) &	  &1 & 2 & 4  & 8\\ \shline
		Binomial	 &S & 71.9& 80.0& 86.4  &91.0    \\  
		LiftedStruct$^*$&N &  69.7&  79.3&  86.2  & 91.0  \\ 
		MS mining		&P  & 67.0& 77.4& 84.7& 90.0\\ \hline
		BinLifted   &SN & 70.4& 79.5& 86.2& 91.1 \\ 
		MS weighting&SN & 73.2& 81.5& 87.6& 92.6 \\ \hline
		Binomial$_{m}$ & SP & 74.6& 83.8& 89.7& 94.1\\
		LiftedStruct$_{m}^{*}$ & NP& 72.2& 81.7& 88.0& 92.4\\ \hline
		MS 		&SNP & \bf 77.3& \bf 85.3& \bf 90.5& \bf 94.2 \\ 
	\end{tabular}
	\vspace{1pt}
	\caption{Results on Cars-196 using \textbf{different pair-based methods}. The first column lists the methods and the types of similarities considered in their weighting. Embedding size is set to 64. Subscript ${m}$ denotes applying our MS mining step before weighting.} 
	\label{tab-ablation}
	\vspace{-10pt}
\end{table}

\subsection{Ablation Study}
\label{ablation}

To  demonstrate the importance of weighting the pairs from multi-similarities, we conduct an ablation study on Cars-196 and results are shown in Table \ref{tab-ablation}.
We describe LiftedStruct$^*$, MS mining and MS weighting here, other methods have already been mentioned in section \ref{section-weighting}.  

{\bf{LiftedStruct$^*$.} }   Lifted structure loss is easy to get stuck in a local optima, resulting in poor performance. To evaluate the efficiency of three similarities more clearly and fairly, we make a minor modification to the lifted structure loss, allowing it to employ {\it Similarity-N} more effectively:
\begin{equation}
\small
\label{eq-lift*}
\mathcal{L}_{lift*} = \frac{1}{m}\sum_{i=1}^m  \bigg\{\frac{1}{\alpha}  { \log  \sum_{\vy_k = \vy_i } e^{-\alpha S_{ik}}}  + 
 \frac{1}{\beta}  { \log \sum_{\vy_k \neq \vy_i}
	e^{\beta S_{ik}}} \bigg\}, 
\end{equation} 
where $\alpha=2$, $\beta=50$. This modification is motivated to make lift structure loss more focus on informative pairs, especially the hard negative pairs, and allows it to get rid of the side effect of enormous easy negative pairs. We found that this modification can boost the performance of lifted-structure loss empirically, e.g., with an over 20\%  improvement of Recall@1 on the CUB200 .

{\bf MS mining.} To investigate the impact of each component of MS Loss, we evaluate the performance of MS mining individually, where the pairs selected into $\mathcal{N}_i$ and $\mathcal{P}_i$ are assigned with an equal weight.

{\bf MS weighting.} Similarly, MS weighting scheme is also evaluated individually without the mining step in the ablation study, allowing us to analyze the effect of each similarity more perspicaciously. In MS weighting, each pair in a mini-batch is assigned with a non-zero weight, according to the weighting strategy described in Eqn \ref{eq-neg} and Eqn \ref{eq-pos}.
 
With the performance reported in Table \ref{tab-ablation}, we analyze the effect of each similarity as below:

{\it Similarity-S:} A cosine self-similarity is of the most importance. Binomial deviance loss, based on the {\it Similarity-S}, achieves the best performance by using a single similarity. Moreover, our MS weighting outperforms LiftedStruct$_m^*$ by $69.7\% \rightarrow 73.2\%$ at recall@1, and  Binomial$_m$ also improves the recall@1 with $67.0\% \rightarrow74.6\%$ over the MS mining,  by adding the {\it Similarity-S} into their weighting schemes. 

{\it Similarity-N:} Relative similarities are also helpful to measuring the importance of a pair more precisely. With {\it Similarity-N}, our MS weighting increases the Recall@1 by 1.3\% over Binomial ($71.9\% \rightarrow73.2\%$). Moreover, with {\it Similarity-N}, LiftedStruct$_m^*$ obtains a significant performance improvement over MS sampling ($ 67\%\rightarrow72.2\%$), by considering both {\it Similarity-P} and {\it Similarity-N}.

{\it Similarity-P:} As shown in Table \ref{tab-ablation}, by adding a mining step based on {\it Similarity-P}, the performances of LiftedStruct$^*$, Binomial and MS weighting are consistently improved by a large margin. For instance, Recall@1 of Binomial is increased by nearly 3\%: $71.9\% \rightarrow 74.6\%$.  

Finally, the proposed MS loss achieves the best performance among these methods, by exploring multi-similarities for pair mining and weighting. However, it is critical to integrate the three similarities effectively into a single framework where the three similarities can be fully explored and optimized jointly by using a single loss function. For example, BinLifted, which uses a weighting scheme considering both {\it similarities-S} and {\it similarities-N}, has lower performance than that of single Binomial, since it implements a simple and straightforward combination of Binomial and LiftedStruct$_m^*$. More discussions on the difference between our MS weighting and the direct combination are presented in Supplementary Material.

\begin{table*}[t]
    \tablestyle{10pt}{1.1}
	\begin{center}
		\begin{tabular}{l|cccccc|cccccc}
			& \multicolumn{6}{c}{CUB-200-2011}  &\multicolumn{6}{c}{Cars-196}\\
			Recall$@K$  (\%) & 1 & 2 & 4 & 8 & 16 & 32 & 1 & 2 & 4 & 8 & 16 & 32\\ \shline
			{Clustering $^{64}$\cite{struct-clustering}}  &48.2 & 61.4 & 71.8 & 81.9 & - & - & 58.1 & 70.6 & 80.3 & 87.8 & - & -\\ 
			{ProxyNCA$^{64}$ \cite{proxyloss}} &49.2 & 61.9 & 67.9 & 72.4 & - & - & 73.2 & 82.4 & 86.4 & 87.8 & - & -\\
			{Smart Mining$^{64}$ \cite{smart-mining}} &49.8 & 62.3 & 74.1 & 83.3 & - & - & 64.7 & 76.2 & 84.2 & 90.2 & - & -\\ 
			{Margin$^{128}$ \cite{sampling}}& 63.6& 74.4& 83.1& 90.0& 94.2 & - & 79.6& 86.5& 91.9& 95.1& 97.3 & - \\
			HDC$^{384}$ \cite{struct-clustering}& 53.6 & 65.7 & 77.0 & 85.6 & 91.5 & 95.5 & 73.7 & 83.2 & 89.5 & 93.8 & 96.7 & 98.4\\ 
			HTL$^{512}$ \cite{HTL} & 57.1& 68.8& 78.7& 86.5& 92.5& 95.5 & 81.4& 88.0& 92.7& 95.7& 97.4& \bf{99.0} \\\hline
			ABIER$^{512}$ \cite{bier}&57.5 &68.7 &78.3 &86.2 &91.9 &95.5 &82.0 &89.0 &93.2 &96.1 &97.8 &98.7\\
			ABE$^{512}$ \cite{Kim_2018_ECCV}  & 60.6 & 71.5 & 79.8 & 87.4 & - & - & \bf 85.2 & \textbf{90.5} & \textbf{94.0} & 96.1 & - & -\\
			\hline 
			MS$^{64}$& 57.4& 69.8&  80.0&  87.8& 93.2& 96.4 & 77.3& 85.3& 90.5& 94.2& 96.9& 98.2\\ 
			MS$^{512}$& \bf65.7& \bf77.0& \bf86.3& \bf91.2& \bf95.0& \bf97.3& 84.1& 90.4 & \bf94.0& \bf96.5& \bf98.0& 98.9\\
		\end{tabular}
		\vspace{3pt}
		\caption{\textbf{Recall@$K(\%)$ performance on CUB200 and Cars-196.} Superscript denotes embedding size. ABIER \cite{bier} and ABE \cite{Kim_2018_ECCV} are ensemble methods.}
		\label{cub-car-table}
		\vspace{-15pt}
	\end{center}
\end{table*}
\subsection{On Embedding Size}
\begin{figure}[htp]
	\vspace{-10pt}
	\centering
	\includegraphics[width=0.5\textwidth]{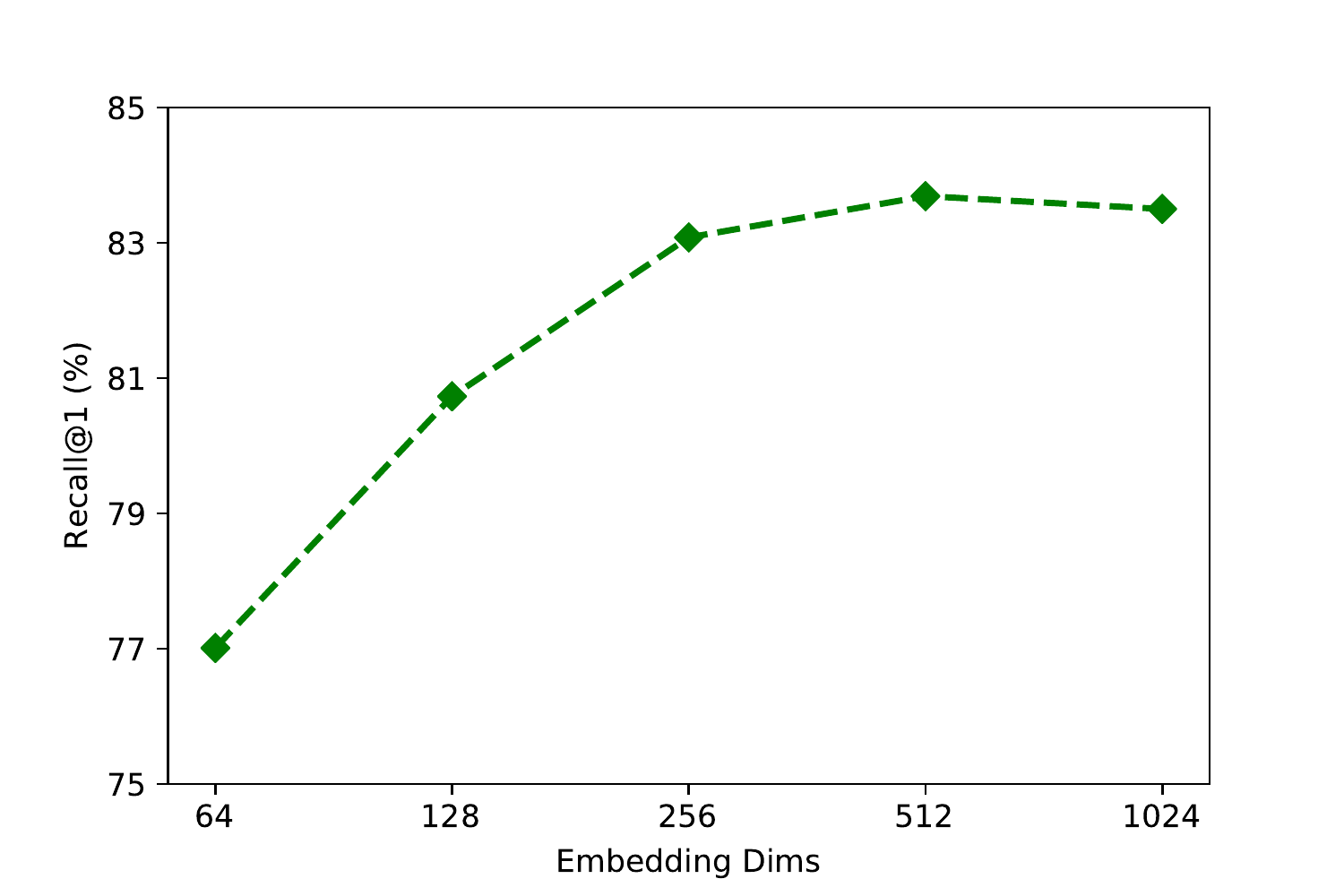}
	\caption{Results on the Cars-196 dataset using \textbf{different embedding sizes} of our MS loss. Larger embedding size can considerably improve recall@1, while embedding sizes larger than 512 are not necessary.}
	\label{fig-embeding-size}
\end{figure} 
By following \cite{semi-hard}, we study the performance of MS loss with varying embedding sizes $\{64, 128, 256, 512, 1024\}$. As shown in Figure \ref{fig-embeding-size}, the performance is increased consistently with the embedding dimension except at  1024. This is different from lifted structure loss, which achieves its best performance at 64 on the Cars-196 dataset  \cite{lifted-structured-loss}. 

\begin{table}[t]
    \tablestyle{7.5pt}{1.1}
	\begin{center}
		\begin{tabular}{l|cccccc}
			Recall$@K$ (\%) & 1 & 10 & 20 & 30 & 40 & 50\\ \shline
			FashionNet$^{4096}$ \cite{DeepFashion}&53.0 & 73.0 & 76.0 & 77.0 & 79.0 & 80.0 \\
			HDC$^{384}$ \cite{struct-clustering}& 62.1 & 84.9 & 89.0 & 91.2 & 92.3 & 93.1\\	 
			HTL$^{128}$ \cite{HTL} & 80.9& 94.3& 95.8& 97.2& 97.4& 97.8\\\hline
			ABIER$^{512}$ \cite{bier}  & 83.1 & 95.1 & 96.9 & 97.5 & 97.8 & 98.0\\	
			ABE$^{512}$ \cite{Kim_2018_ECCV} & 87.3 & 96.7 & 97.9 & 98.2 & 98.5 & 98.7\\ 
			\hline
			{MS}$^{128}$  & 88.0& 97.2& 98.1& 98.5& 98.7& 98.8\\
			{MS}$^{512}$   &\bf89.7 &\bf97.9& \bf98.5& \bf98.8 & \bf99.1 & \bf99.2\\
		\end{tabular}
		\caption{\textbf{Recall@$K(\%)$ performance on In-Shop.}}
		\label{Shop-table}
		\vspace{-10pt}
	\end{center}
\end{table}

\subsection{Comparison with State-of-the-Art}
\begin{table}[htp]
    \tablestyle{12pt}{1.1}
	\begin{center}
		\begin{tabular}{l|cccc}
			Recall$@K$ (\%)& 1 & 10 & 100 & 1000\\ \shline
			{Clustering $^{64}$\cite{struct-clustering}}  & 67.0 & 83.7 & 93.2 & -\\
			{ProxyNCA$^{64}$} \cite{proxyloss} & 73.7 & -& -& -\\
			{Margin$^{128}$ \cite{sampling}} & 72.7 & 86.2 & 93.8 & 98.0\\
			HDC$^{384}$ \cite{struct-clustering}& 69.5 & 84.4 & 92.8 & 97.7 \\	
			HTL $^{512}$ \cite{HTL} & 74.8& 88.3& 94.8& 98.4\\\hline
			ABIER$^{512}$ \cite{bier}  & 74.2 & 86.9 & 94.0 & 97.8 \\	
			ABE$^{512}$ \cite{Kim_2018_ECCV} &76.3 & 88.4 & 94.8 & 98.2\\ 
			\hline
			{MS}$^{64}$   & 74.1& 87.8& 94.7& 98.2 \\
			{MS}$^{128}$   & 76.6& 89.2& 95.2& 98.4\\
			{MS}$^{512}$   & \bf  78.2& \bf  90.5& \bf  96.0& \bf  98.7 \\
		\end{tabular}
		\vspace{4pt}
		\caption{\textbf{Recall@$K(\%)$ performance on SOP.}}
		\label{sop-table}
		\vspace{-10pt}
	\end{center}
\end{table}

We further compare the performance of our method with the state-of-the-art techniques on image retrieval task.
As shown in Table \ref{cub-car-table}, our MS loss improves Recall@1 by 7\%  on the CUB200, and 4\% on the Cars-196 over Proxy-NCA at dimension 64.  Compared with ABE employing an embedding size of 512 and attention modules, our MS loss achieves a higher Recall@1 by +5\% improvement at the same dimension on the CUB200.
On the Cars-196, our MS loss obtains the second best performance in terms of Recall@1, while the best results are achieved by ABE, which applies an ensemble method with a much heavier model. Moreover, on the remaining three datasets, our MS loss is of much stronger performance than ABE. 

In Table \ref{Shop-table} and \ref{sop-table}, our MS loss outperforms Proxy-NCA by 0.4\% and Clustering by 7\% at the same embedding size of 64. Furthermore, when compared with ABE, MS loss increases Recall@1 by 1.9\%  and  2.7\% on the SOP and In-Shop respectively. Moreover, even with much compact embedding features of 128 dimension, our MS loss has already surpassed all existing methods, which utilize much higher dimensions of 384, 512 and 4096.

To summarize, on both fine-grained datasets like Cars-196 and CUB200, and large-scale datasets with enormous categories like SOP and In-Shop, our method achieves new state-of-the-art or comparable performance, even taking those methods with ensemble techniques like ABE and BIER into consideration.  

\section{Conclusion}
We have established a General Pair Weighting (GPW) framework and presented a new multi-similarity loss to fully exploit the information of each pair. Our GPW, for the first time, unifies existing pair-based metric learning approaches into general pair weighting through gradient analysis. It provides a powerful tool for understanding and explaining various pair-based loss functions, which allows us to clearly identify the main differences and key limitations of existing methods. Furthermore, we proposed a multi-similarity loss which considers all three similarities simultaneously, and developed an iterative pair mining and weighting scheme for optimizing the multi-similarity loss efficiently. Our method obtains new state-of-the-art performance on a number of image retrieval benchmarks.
{\small
	\bibliographystyle{ieee}
	\bibliography{egbib}
}

\clearpage
\clearpage


\section*{Supplementary material}

# Multi-Similarity Loss with GPW for Deep Metric Learning
# Supplementary Material

## 1. Introduction

This supplementary material provides more details of General Pair Weighting (GPW) framework and Multi-Similarity (MS) loss. First, we further revisit three existing pair-based loss functions under our GPW framework. Then, we analyze the limitation of direct combination of binomial deviance loss [6] and lifted structure loss [1] (referred as BinLifted), compared to our weighting scheme. Finally, we analyze the impact of batch-size to the proposed MS loss with experiments conducted on CUB200 and SOP datasets.

## 2. Revisit Pair-based Loss Functions

In this section, we further analyze three recent pair-based loss functions in addition to those presented in the main paper: N-pairs loss [3], NCA loss [2] and histogram loss [4].

**N-pairs Loss.** Sohn *et al.* [3] proposed a N-pairs loss, which is a special case of the lifted structure loss by only considering single positive pair. It can be analyzed by following the same process of analyzing the lifted structure loss in the main paper, as shown in Eq. (6)-(8).

**NCA Loss.** Salakhutdinov *et al.* [2] introduced a NCA loss to learn a nonlinear embedding to optimize the classification performance of a soft-KNN classifier:

$$\mathcal{L}_{nca} := \sum_{i=1}^{m} log \frac{\sum_{\boldsymbol{y}_k = \boldsymbol{y}_i} e^{S_{ik}}}{\sum_{i=1}^{m} e^{S_{ik}}}$$
$$= \sum_{i=1}^{m} \left[ log \sum_{\boldsymbol{y}_k = \boldsymbol{y}_i} e^{S_{ik}} - log \sum_{i=1}^{m} e^{S_{ik}} \right]. \quad (1)$$

By following GPW framework, the weight of a pair $\{\boldsymbol{x}_i, \boldsymbol{x}_j\}$, *i.e.*, $w_{ij}$, can be derived from differentiating $\mathcal{L}_{nca}$ with respect to $S_{ij}$:

$$w_{ij}^{+} = \frac{e^{S_{ij}}}{\sum_{\boldsymbol{y}_k = \boldsymbol{y}_i} e^{S_{ik}}} - \frac{e^{S_{ij}}}{\sum_{i=1}^{m} e^{S_{ik}}}$$
$$= \frac{1}{\sum_{\boldsymbol{y}_k = \boldsymbol{y}_i} e^{S_{ik} - S_{ij}}} - \frac{1}{\sum_{i=1}^{m} e^{S_{ik} - S_{ij}}}, \quad (2)$$
$$w_{ij}^{-} = \frac{e^{S_{ij}}}{\sum_{i=1}^{m} e^{S_{ik}}} = \frac{1}{\sum_{i=1}^{m} e^{S_{ik} - S_{ij}}}.$$

As can be found in Eq. 2, the weight for a positive pair or a negative pair is calculated based on its relative similarities, which compute a similarity between a current pair and its neighboring pairs. The NCA loss focuses on hard negative pairs and the high-confident positive pairs, collected from the neighboring pairs of an anchor point, in the embedding space.

**Histogram Loss.** Ustinova *et al.* [4] designed a histogram loss based on quadruplets, which can be formulated as:

$$\mathcal{L}_{hist} := \sum_{r=1}^{R}(h_r^{-} \sum_{q=1}^{r} h_q^{+})$$
$$= \sum_{r=1}^{R}(\sum_{q=1}^{r} h_q^{+}) h_r^{-} \quad (3)$$
$$= \sum_{q=1}^{R}(\sum_{r=q}^{R} h_r^{-}) h_q^{+}$$

where $R$ is the dimension of a histogram for positive or negative cosine similarities. $h_q^{+}$ is the histogram estimation at node $q$ of the cosine similarity for a positive pair, and $h_r^{-}$ is that of a negative pair at node $r$.

$$h_r^{+} = \frac{1}{|\mathcal{S}^{+}|} \sum_{\boldsymbol{y}_i = \boldsymbol{y}_j} \delta_{ijr}, \quad (4)$$

where $\delta_{ijr}$ is defined as:

$$\delta_{ijr} = \begin{cases} (S_{ij} - t_{r-1})/\Delta, & S_{ij} \in [t_{r-1}, t_r], \\ (t_{r+1} - S_{ij})/\Delta, & S_{ij} \in [t_r, t_{r+1}], \\ 0, & otherwise, \end{cases} \quad (5)$$

where $\Delta = 2/(R-1)$, $t_r = r\Delta - 1$. Estimation of $h_q^{-}$ is computed analogously.

To better understanding histogram loss within our GPW framework, we first provide the following formulations (more details can be found in [4]),

1

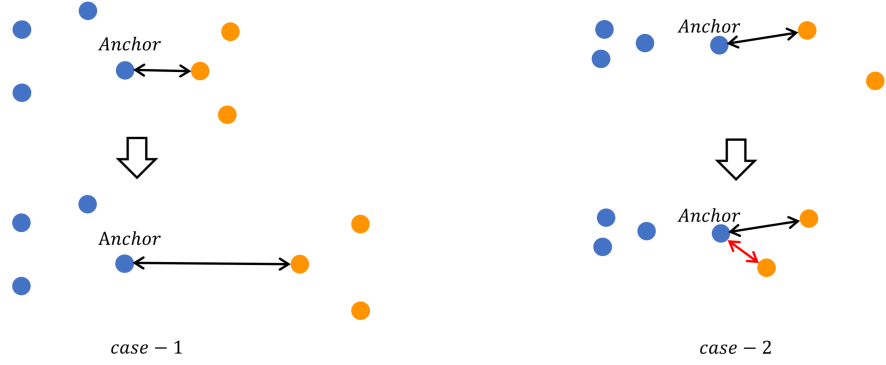

Figure 1. Limitations of the BinLifted weighting scheme: (i) in case-1, the negative pair on the bottom, which is of much lower cosine similarity than the top one, will be assigned with a larger weight; (ii) in case-2, a negative pair is fixed when the neighboring negative samples move closer to the anchor, which reduce the relative similarly (*Similarity-N*) of the negative pair.

$$\frac{\partial \mathcal{L}_{hist}}{\partial h_q^+} = \sum_{r=q}^{R} h_r^-$$
$$\frac{\partial \mathcal{L}_{hist}}{\partial h_r^-} = \sum_{q=1}^{r} h_q^+ \quad (6)$$

For a positive pair $S_{ij}$

$$\frac{\partial h_r^+}{\partial S_{ij}} = \begin{cases} \frac{+1}{\Delta|\mathcal{S}^+|}, & S_{ij} \in [t_{r-1}, t_r], \\ \frac{-1}{\Delta|\mathcal{S}^+|}, & S_{ij} \in [t_r, t_{r+1}], \\ 0, & otherwise, \end{cases} \quad (7)$$

where $|\mathcal{S}^+|$ is the number of positive pairs. Then we have

$$\frac{\partial h_r^-}{\partial S_{ij}} = 0, \quad (8)$$

since $h_r^-$ is calculated from negative pairs, and thus is unrelated to the similarities of positive pairs, $S_{ij}$.

Finally, the partial derivative of $\mathcal{L}_{hist}$ w.r.t. the similarity of a positive pair $S_{ij} \in [t_p, t_{p+1}]$ is:

$$\begin{aligned} \frac{\partial \mathcal{L}_{hist}}{\partial S_{ij}} &= \sum_{q=1}^{R} \frac{\partial \mathcal{L}_{hist}}{\partial h_q^+} \frac{\partial h_q^+}{\partial S_{ij}} + \sum_{r=1}^{R} \frac{\partial \mathcal{L}_{hist}}{\partial h_r^-} \frac{\partial h_r^-}{\partial S_{ij}} \\ &= \sum_{q=1}^{R} \left( \sum_{r=q}^{R} h_r^- \right) \frac{\partial h_q^+}{\partial S_{ij}} \\ &= \left( \sum_{r=p}^{R} h_r^- \right) \frac{\partial h_p^+}{\partial S_{ij}} + \left( \sum_{r=p+1}^{R} h_r^- \right) \frac{\partial h_{p+1}^+}{\partial S_{ij}} \\ &= \frac{1}{\Delta|\mathcal{S}^+|} \left( \sum_{r=p+1}^{R} h_r^- - \sum_{r=p}^{R} h_r^- \right) \\ &= \frac{-1}{\Delta|\mathcal{S}^+|} h_p^-. \end{aligned} \quad (9)$$

Thus the weight value assigned to this positive pair is $\frac{1}{\Delta|\mathcal{S}^+|} h_p^-$. Similarly, for a negative pair with a cosine similarity of $S_{ij} \in [t_p, t_{p+1}]$, the weight under a histogram loss is $\frac{1}{\Delta|\mathcal{S}^-|} h_{p+1}^+$.

Though with complicated formulation and rough derivation, pair weight scheme of histogram loss can be presented in an extremely concise and clear form, as shown in Eq. 9. $h_p^-$ is approximately the ratio of negative pairs which have a lower cosine similarity compared with the current positive pair, and $\frac{1}{\Delta|\mathcal{S}^-|}$ can be regarded as a fixed normalizer. Similarly, the weight of a negative pair is the ratio of positive pairs having a lower cosine similarity than the current negative one. Therefore, the weighting scheme clearly indicates that a histogram loss estimates pair weights only based on *Similarity-P*, e.g., by comparing the negative pairs with the positive ones, and vise verse. This may reduce the performance of histogram loss, leading to lower performance than that of binomial deviance loss on the CUB200 and SOP, as reported in [4].

## 3. BinLifted v.s. MS Weighting

As discussed in the main paper, our MS loss is related to a direct combination of binomial deviance loss and lifted structure loss (referred as BinLifted), by jointly considering both *Similarity-S* and *Similarity-N*. In our ablation study presented in Section 5.1 (in main paper), we show by experiments that our weighting scheme is superior to that of BinLifted. Here we explain the benefits of our iterative mining and weighting scheme. Given an example of a negative pair $\{x_i, x_j\}$, the weight computed by Binlifted weighting scheme $\widehat{w}_{ij}$ is:

$$\widehat{w}_{ij} = \frac{1}{2} \left( \frac{e^{\beta(S_{ij}-\lambda)}}{1+e^{\beta(S_{ij}-\lambda)}} + \frac{e^{\beta S_{ij}}}{\sum_{y_k \neq y_i} e^{\beta S_{ik}}} \right). \quad (10)$$

From Eq. 10, the weight of BinLifted satisfies:

$$\widehat{w}_{ij} > \frac{1}{2} \max \left( \frac{e^{\beta(S_{ij}-\lambda)}}{1+e^{\beta(S_{ij}-\lambda)}}, \frac{e^{\beta S_{ij}}}{\sum_{y_k \neq y_i} e^{\beta S_{ik}}} \right). \quad (11)$$

which provides a lower bound for the weighting by BinLifted method. We found that a negative pair with a large relative similarity, will have a weight of $\frac{e^{\beta S_{ij}}}{\sum_{y_k \neq y_i} e^{\beta S_{ik}}}$ which is close to 1, so that BinLifted method will assign a large weight to this pair, and ignore or reduce the impact from *Similarity-S*. As shown in case-1 of Fig. 1, a negative pair on the bottom has a much lower *Similarity-S* than the negative pair on the top, but BinLifted will compute a similar weight for the two pairs. Furthermore, when the *Similarity-S* of a pair is higher than a threshold $\lambda$, the weight will be computed as $\frac{e^{\beta(S_{ij}-\lambda)}}{1+e^{\beta(S_{ij}-\lambda)}}$, which can be a large value, regardless of *Similarity-N*. As illustrated in case-2 of Fig. 1, two negative pairs on the bottom will be assigned with a similar weight, due to large values of *Similarity-S*, while omitting the large difference on the relative similarities (*Similarity-N*) between the two pairs. However, a negative pair on the top of case-2 will also be assigned with a similar weight to the bottom one, even its *Similarity-S* is much larger.

In summary, BinLifted estimates the weight of a pair by mainly considering the larger value computed from *Similarity-S* or *Similarity-N*, while neglecting the smaller one. This drawback may reduce the performance considerably, which can be even lower than that of using a single binomial deviance loss, as shown in the ablation study.

In contrast, our MS weighting is able to compute the weight for a negative pair dynamically:

$$w_{ij} = \frac{e^{\beta(S_{ij}-\lambda)}}{1+\sum_{y_k \neq y_i} e^{\beta(S_{ik}-\lambda)}}. \qquad (12)$$

which addresses the issue raised in case-1 and case-2 of Fig. 1. This allows it to explore more meaningful information from both *Similarity-S* and *Similarity-S* (Eq. 12), rather than only focusing on the single one with the larger value. Positive pairs can be analyzed analogously.

## 4. Effect of Batch Size

To analyze the performance of our MS loss with different batch-size, we conduct experiments on the SOP [1] and CUB200 [5] datasets. We set the embedding size to 512 and $K = 5$, and use Adam optimizer with a learning rate of $10^{-5}$ for all experiments. The recall@1 performance of MS loss on the CUB200 and the SOP are reported in Tables 1 and 2, with batch-size set at $\{20, 40, 80, 160, 240\}$ and $\{20, 40, 80, 160, 320, 640, 1000\}$ respectively.

We observed that batch-size effects the performance of MS loss differently on the CUB200 and SOP: (i) the CUB200 is less sensitive to the change of batch-size than the SOP; (ii) the performance on the CUB200 decreases when the batch-size is increased, while the performance on the SOP benefits from large batch-sizes significantly.

We found that the batch-size impacts a dataset with large inter-class variations more significantly than the small ones.

The CUB200 is a fine-grained dataset with smaller inter-class variations than the SOP, making it easier to collect hard negative pairs with subtle difference. However, by using a small batch-size of 20 on the SOP dataset, our MS mining is difficult to collect any informative pair at some iterations, which happened at over 20% iterations in our experiments. Therefore, a large batch-size, *e.g.*, $> 320$, is required for a dataset with large variations like the SOP. This ensures it to collect enough hard negative pairs, which are meaningful to train a discriminative model, and thus improve the performance.

| Batch-size | Recall@1 (%) |
|---|---|
| 20 | 64.75 |
| 40 | 65.07 |
| 80 | **65.65** |
| 160 | 65.43 |
| 240 | 64.60 |

Table 1. Recall@1 performance of MS loss at the batch-size of $\{20, 40, 80, 120, 160, 240\}$ on the CUB200 database.

| Batch-size | Recall@1 (%) |
|---|---|
| 20 | 71.40 |
| 40 | 73.82 |
| 80 | 75.61 |
| 160 | 76.63 |
| 320 | 77.59 |
| 640 | 78.19 |
| 1000 | **78.35** |

Table 2. Recall@1 performance of MS loss at the batch-size of $\{20, 40, 80, 160, 320, 640\}$ on the SOP database.



\end{document}